\documentclass[aps,pre,twocolumn,superscriptaddress,showpacs]{revtex4}
\usepackage{amssymb,amsmath,latexsym}
\usepackage{epsfig}
\usepackage{graphicx}
\usepackage{color}

\begin{document}

\title{Modeling the average shortest path length in growth of word-adjacency networks}

\author{Andrzej Kulig}
\affiliation{Institute of Nuclear Physics, Polish Academy of Sciences, Krak\'ow, Poland}
\author{Stanis{\l}aw Dro\.zd\.z}
\affiliation{Institute of Nuclear Physics, Polish Academy of Sciences, Krak\'ow, Poland}
\affiliation{Faculty of Physics, Mathematics and Computer Science, Cracow University of Technology, Krak\'ow, Poland}
\author{Jaros{\l}aw Kwapie\'n}
\author{Pawe{\l} O\'swi\c ecimka}
\affiliation{Institute of Nuclear Physics, Polish Academy of Sciences, Krak\'ow, Poland}

\date{\today}

\begin{abstract}

We investigate properties of evolving linguistic networks defined by the word-adjacency relation. Such networks belong to the category of networks with accelerated growth but their shortest path length appears to reveal the network size dependence of different functional form than the ones known so far. We thus compare the networks created from literary texts with their artificial substitutes based on different variants of the Dorogovtsev-Mendes model and observe that none of them is able to properly simulate the novel asymptotics of the shortest path length. Then, we identify the local chain-like linear growth induced by grammar and style as a missing element in this model and extend it by incorporating such effects. It is in this way that a satisfactory agreement with the empirical result is obtained.

\end{abstract}


\maketitle

\section{Introduction}

Essential features of many real-world systems can be expressed by networks with growing number of nodes and edges. Hardware structure of the Internet~\cite{adamic2002,zhou2004,zhang2008}, its information content consisting of linked WWW documents~\cite{albert1999,huberman1999,krapivsky2002,smith2007}, social systems~\cite{jin2001,catanzaro2004,capocci2006}, user-object systems~\cite{zhou2007}, collaborations~\cite{newman2001,guimera2005,tessone2011}, scientific paper citations~\cite{bilke2001,lehmann2005,krapivsky2005}, and epidemic networks~\cite{satorras2001}, to list a few, are systems, in which continuous inflow of new elements and relations leads to substantial system growth.

A system that is the central subject of this work $-$ natural language $-$ can also be represented by growing networks, both on the global and local scales. By treating individual words as the basic constituents of any message transmitted via language, one may define various relations among the words based on their positions in this message, function, or meaning, and construct a related network with the words as its nodes. From a local perspective, the growth of such a network can be realized by expanding the message (e.g., writing a piece of text) by adding to the already existing part of it both the new words that were not used there before, and repeating the old words put in new contexts. Globally, by considering the network formed from a giant corpus consisting of all the existing written and spoken text samples, new nodes may be identified with newly coined words that appear from time to time in every natural language.

As the succession of words in a piece of text reflects both the common and individual properties of natural language, including grammar and author's style, the word-adjacency networks seem to be a very interesting example of linguistic networks. They are constructed from text samples after linking words that are direct neighbours of each other at least once in a sample~\cite{ferrer2001,masucci2006}. On the statistical level, such networks are expected to inherit selected properties of the word frequencies described by the Zipf-Mandelbrot law~\cite{zipf1949} (or its double-scaling version~\cite{montemurro2001,ferrer2001}), the Heaps law~\cite{herdan1960,heaps1978}, as well as certain grammar and stylistic rules, and other properties.

Grammar can influence both the local and global properties of linguistic networks. Locally it can lead to correlations or anticorrelations in word usage; this is the obvious action of grammar. On the other hand, it acts on global scale as well, as it can develop hubs corresponding to the words that play purely grammatical roles in sentences (articles, prepositions, conjunctions) and thus influencing the overall network topology. However, the latter effect mingles with the statistical properties of language, so one may effectively restrict the role of grammar to local scales only.

It is not a surprise that the empirical word-adjacency networks constructed from text samples like novels or scientific publications exhibit strong hierarchical structure with hubs $-$ the nodes of large connectivity that correspond to words with the highest frequency of use $-$ and peripheral nodes, which are linked to few neigbours and correspond to rarely used words. These nodes are connected among themselves in highly disassortative manner, i.e., the hubs usually form connections with the peripheral nodes and not with other hubs~\cite{ferrer2001}. The connectivity distribution of nodes shows scale-free behaviour with the power exponent $2 < \alpha < 3$, or, if the samples are large enough to obtain networks with sufficient number of nodes, the connectivity distributions show another power-law regime with $\beta > 3$, this can be related to two such regimes in the word frequency rank-plots~\cite{montemurro2001,ferrer2001}.

The so-defined linguistic networks can in principle be modelled in three different ways. (1) By defining a stochastic process that can mimic the process of text creation and, then, to form the corresponding network in the same manner as the empirical networks are built. In this context, one may exploit various derivatives of the Yule-Simon processes~\cite{yule1925,simon1955}. (2) By constructing an explicit growing network model, in which an initial minimal network core is expanded by adding new nodes and new edges according to some predefined rules. Here the most optimal approach uses a class of the accelerated-growth network models, introduced by Dorogovtsev and Mendes (DM)~\cite{dorogovtsev2000a,dorogovtsev2001a,dorogovtsev2001b} and further elaborated by others~\cite{masucci2006,markosova2008}. These models are able to reproduce some characteristics of the empirical word-adjacency networks like the double-scaling connectivity distributions, average connectivity, and clustering coefficients. (3) By considering a random walk on an \textit{a priori} existing primary network and then using the so-obtained sequence of nodes (``words'') as a source for building a secondary network as in (1). Although in such a case the secondary network asymptotically reproduces the primary one, each network may have different properties in early stages of the process.

In this work we study topological properties of empirical word-adjacency networks obtained from literary texts written in different European languages. In particular, we analyze growth of these networks with the stress put on temporal evolution of the average shortest path length (ASPL). In contrast to other network measures like the connectivity distribution or the clustering coefficient, the properties of ASPL in the linguistic networks have yet been rather rarely studied~\cite{grabska2012}. We fill this gap by studying the empirical data and evaluating how well the models agree with the related results from the data.

\section{Empirical data}

We consider networks of adjacent words in written texts. The words are nodes and the edges connect only those words that are adjacent to each other at least once throughout a given text. By a word we mean a unique, transformed to lowercase sequence of characters (letters, digits, and inner hyphens) exactly as it appears between two blanks or punctuation marks, without lemmatization. We ignore full stops and other equivalent punctuation marks in order to avoid the ambiguous situations, in which a network under study forms a few disconnected components between which no path exists. However, a test study based on the largest connected components in sample empirical networks showed that there is no qualitative difference between results of both the approaches. For our analysis, we selected literary texts written in one of the following languages: English, French, German, Polish, Russian, or Spanish. We focus on long novels that comprise 10,000-60,000 distinct words; these numbers also determine the corresponding network sizes.

\begin{figure}
\includegraphics[scale=0.17]{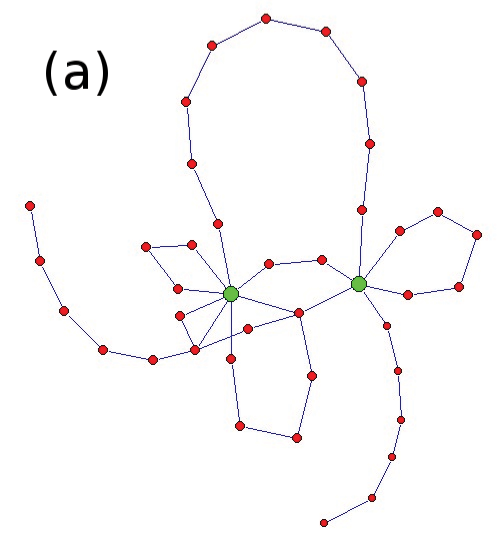}
\includegraphics[scale=0.17]{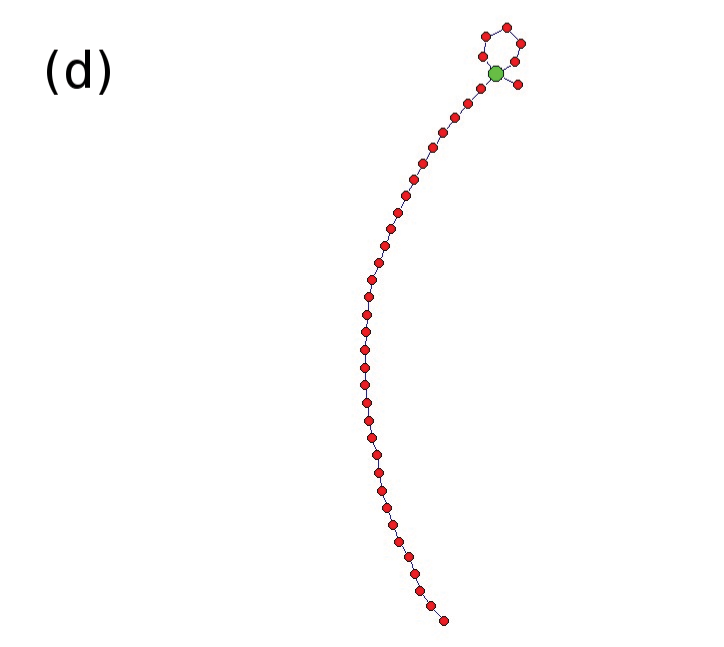}

\includegraphics[scale=0.16]{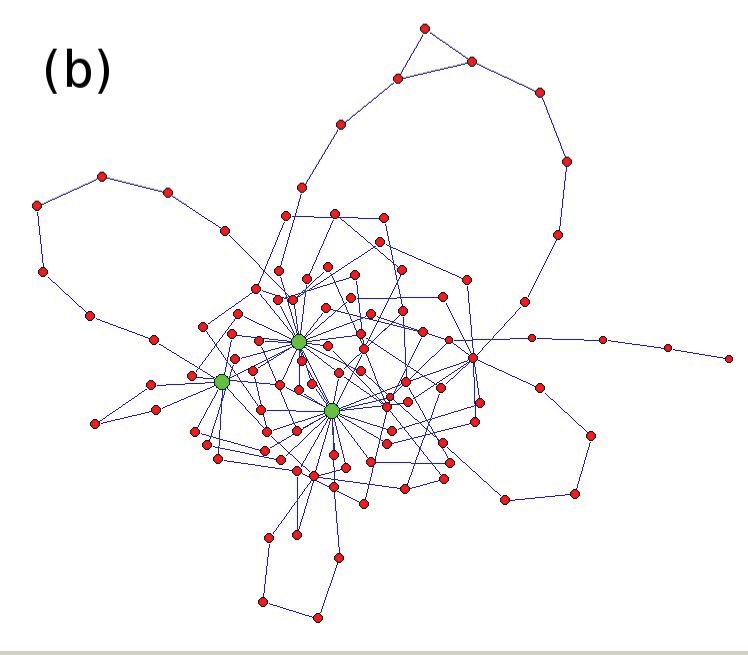}
\includegraphics[scale=0.16]{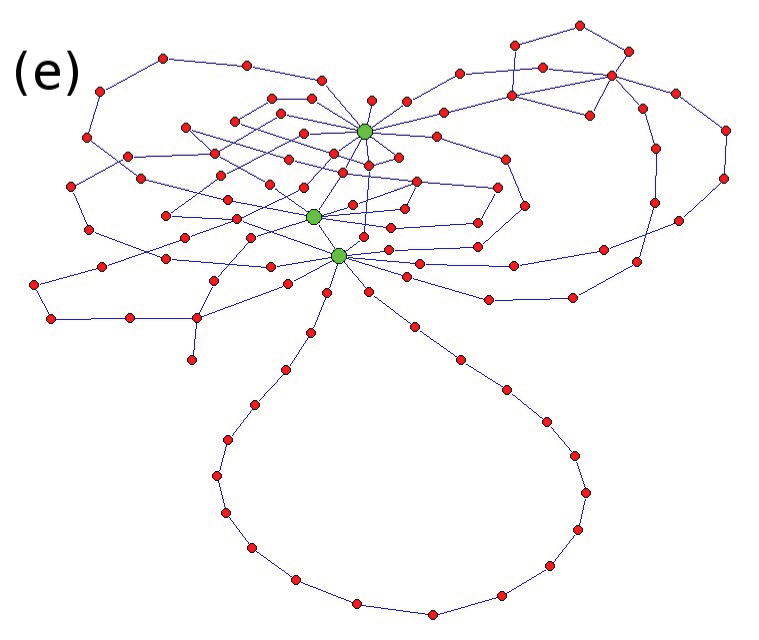}

\includegraphics[scale=0.17]{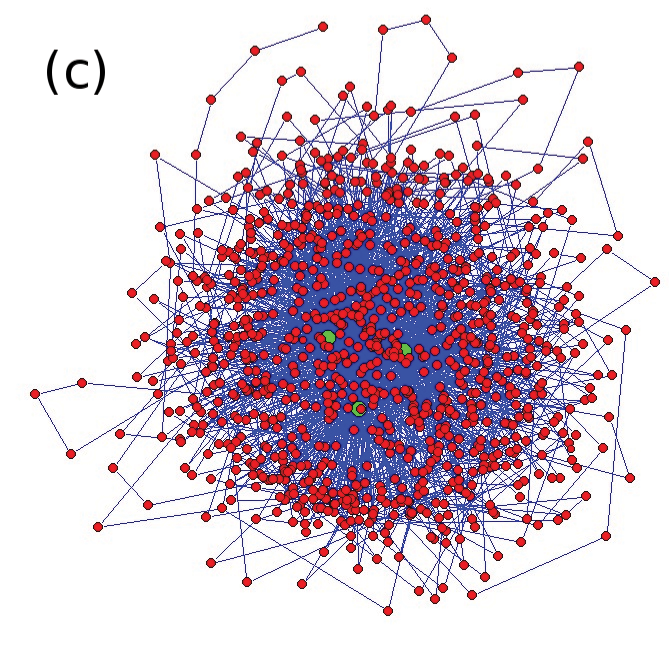}
\includegraphics[scale=0.17]{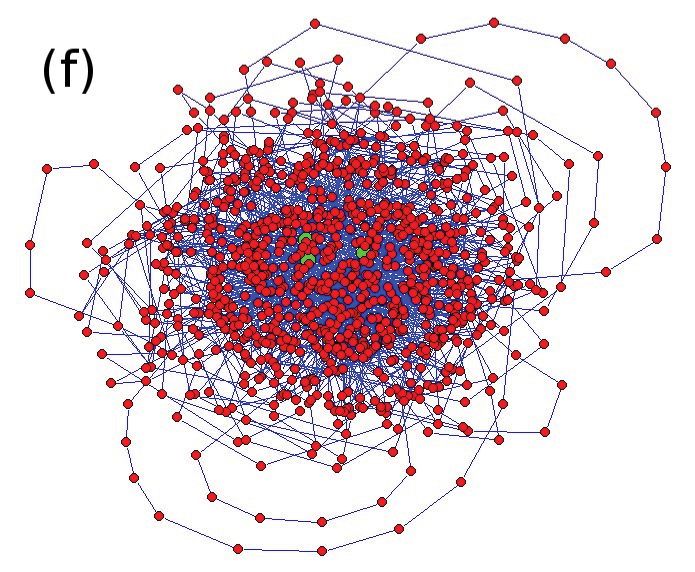}
\caption{(Color online) Growing word-adjacency networks for sample literary texts: \textit{Ulysses} by J.~Joyce (English, left) and \textit{Lalka} by B.~Prus (Polish, right). Each picture represents different stage of the network's development: (a,d) an initial phase with few or even no repeated words ($N=40$), (b,e) a phase in which hubs start to be distinguishable ($N=100$), (c,f) a phase in which old words are used more often than new ones ($N=1000$). The words that play a role of network hubs in (b,e) are distinguished in all panels by larger size and different colour (green/light gray).}
\label{fig.networks}
\end{figure}

Now, let us look at the procedure of real text creation. This procedure is not stationary and consists of at least two phases. Starting from a single word taken (effectively) at random from a dictionary, new words are subsequently added under strict rules of grammar and influence of the planned information content and style. These factors together lead typically to a situation, in which an opening sequence of several unrepeated words form a chain network (see Fig.~\ref{fig.networks}(d)) with the number of edges following the number of nodes: $E(\tau)=N(\tau)-1$, where $\tau$ is the length of a piece of text being created (measured in words). This is an infancy phase of the network's growth. After this phase ends, one of the already-used words occurs repeated and the network can no longer be represented by a pure chain, receiving loops and side branches ($E(\tau) > N(\tau)-1$, Fig.~\ref{fig.networks}(a,b,e)). Despite this, the repetitive use of the old words is still rather limited and new words are frequently added. However, as the text grows further, more and more old words can be exploited again without compromising style and clarity of the message. This leads to a situation that many more new edges linking the existing nodes are added than are the edges that attach new nodes to the existing ones. At this phase the diameter of the network decelerates its increase and the network optically condensates (Fig.~\ref{fig.networks}(c,f)).

Further phases of the network's development crucially depend on the rate of adding new words. Typically, vocabulary (a set of unique words) of real texts grows sublinearly with the text's length, which means that the number of edges $E(\tau)$ increases faster than the number of nodes $N(\tau)$. For moderate values of $N$, this growth approximately obeys the Heaps law stating that vocabulary grows as a power-law function of $\tau$~\cite{herdan1960,heaps1978}:
\begin{equation}
N(\tau) \sim {\tau}^{\delta} , \qquad 0.4 < \delta < 1
\label{eq.heaps}
\end{equation}
(the range of $\delta$ according to Ref.~\cite{bochkarev2009}). In this case, the text growth is more and more dominated by repeating the words that were already in use before, while new words arrive with decreasing frequency. In the network representation, a peculiar situation is possible for sufficiently large $N$: the paths between pairs of nodes tend to decrease with increasing $N$. This property can statistically be expressed by the average shortest path length (ASPL). For binary networks, this quantity is defined as
\begin{equation}
L(N) = {1 \over N(N-1)} \sum_{i,j} d(i,j) ,
\label{average.path}
\end{equation}
where $d(i,j)$ is the shortest path between the nodes $i,j$. The functional character of $L(N)$ crucially depends on the network topology. For equilibrium networks, ASPL is typically an increasing function of the network size with a rate of this increase dependent on the connectivity distribution $P(k)$. For both the classical Erd\"os-R\'enyi (ER) random graphs~\cite{erdoes1960} and the scale-free networks with $\gamma > 3$, in the large network size limit, $L(N) \sim \ln N$; for the BA networks ($\gamma = 3$): $L(N) \sim \ln N / \ln \ln N$. For the fat-tailed networks with $2 < \gamma < 3$, one observes either the ultrasmall-world dependence: $L(N) \sim \ln \ln N$~\cite{cohen2003} or even complete saturation of ASPL at~\cite{fronczak2004}:
\begin{equation}
\lim_{N \to \infty} L(N)={1 \over 2}+{2 \over 3-\gamma} , \qquad 2 < \gamma < 3 .
\label{eq.aspl.fronczak}
\end{equation}

\begin{figure}
\hspace{-0.5cm}
\includegraphics[scale=0.37]{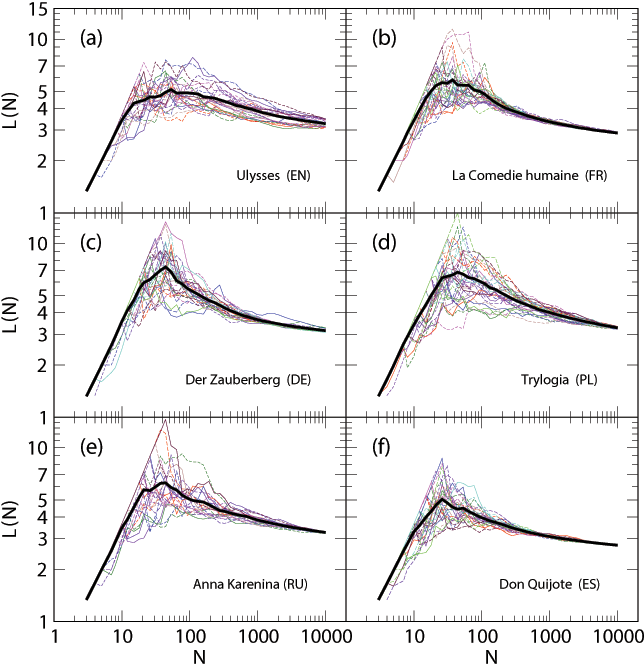}
\caption{(Color online) Evolution of the average shortest path length $L(N)$ for growing word-adjacency networks created from sample literary texts: (a) \textit{Ulysses} by J.~Joyce (written in English), (b) \textit{La Com\'edie humaine} by H.~de~Balzac (French), (c) \textit{Der Zauberberg} by T.~Mann (German), (d) \textit{Trylogia} by H.~Sienkiewicz (Polish), (e) \textit{Anna Karenina} by L.~Tolstoy (Russian), and (f) \textit{Don Quijote} by M.~de~Cervantes (Spanish). Each text was divided into a number of pieces in order to obtain an ensemble of samples. Results for these pieces are shown denoted by different lines in each panel, as well as their average behaviour (thick line).}
\label{fig.texts.aspl}
\end{figure}

\begin{figure}
\hspace{-0.5cm}
\includegraphics[scale=0.37]{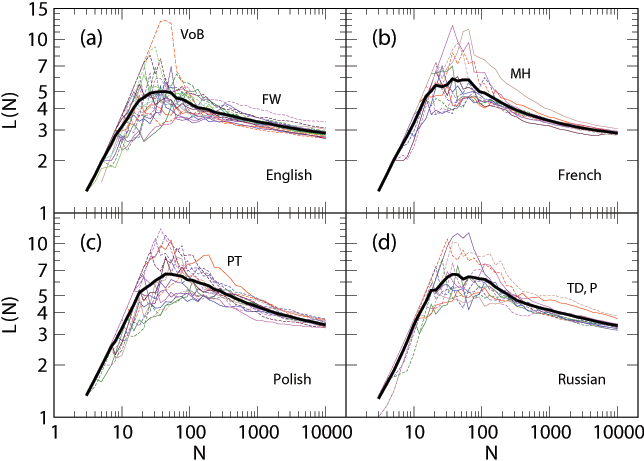}
\caption{(Color online) Evolution of the average shortest path length $L(N)$ for growing word-adjacency networks created from opening pieces of  sample literary texts representing different European languages: (a) English (23 texts), (b) French (14 texts), (c) Polish (17 texts), and (d) Russian (12 texts). Thin lines correspond to individual texts, while thick lines denote average over the texts written in the same language. The texts with peculiar behaviour of $L(N)$ are denoted by acronyms: \textit{VoB} $-$ \textit{The Voyage of the Beagle} by C.~Darwin, \textit{FW} $-$ \textit{Finnegans Wake} by J.~Joyce, \textit{MH} $-$ \textit{M\'emoires d'Hadrien} by M.~Yourcenar, \textit{PT} $-$ \textit{Pan Tadeusz} by A.~Mickiewicz, \textit{TD} $-$ \textit{Tikhiy Don} by M.~Sholokhov, and \textit{P} $-$ \textit{Peterburg} by A.~Bely.}
\label{fig.languages.aspl}
\end{figure}

In this context, Figs.~\ref{fig.texts.aspl} and~\ref{fig.languages.aspl} show how striking is the contrast between these generic model networks and the empirical ones. Initially, for the first added nodes, the size of the world-adjacency networks parametrized by ASPL grows approximately linearly with $N$, then it abruptly switches to the next phase with overall decreasing $L(N)$. This effect can be put in a context of the network's aging: if the network is mature enough, the distances between the nodes tend to assume small values, typically above 2.5 and below 4 for $N=10,000$, depending on a piece of text (Fig.~\ref{fig.languages.aspl}). This is achieved by the increasing frequency of adding intra-networks edges as the texts grow.

At this point two observations have to be stressed. First, for the majority of texts the asymptotic behaviour of $L(N)$ (in practice, for $N \approx 10^4$ or larger) depends little on a sample (Fig.~\ref{fig.texts.aspl}). Even more, if one compares different texts written by different authors who share the same language, it appears that typically the corresponding networks also share their topological properties measured by ASPL. This means that such large-scale properties of text samples express the overall statistical properties of language rather than revealing any individual fingerprints of authors or styles (Fig.~\ref{fig.languages.aspl}). Only the texts that are significantly atypical can develop structure that for large $N$ notably deviates from this common picture. For English literature such an example is doubtlessly the novel \textit{Finnegans Wake} written by J.~Joyce. Indeed, its ASPL assumes the distinctly largest values among the considered English works. The only texts that show comparative distinctness are two Russian novels: \textit{Peterburg} by A.~Bely and \textit{Tikhiy Don} by M. Sholokhov, the latter notable for its exuberant style that might be the origin of their unequally long internode paths. Thick lines in Fig.~\ref{fig.languages.aspl} indicate also that the average values of ASPL are larger for the languages with strong flexion, like Polish and Russian, and smaller for the western European languages, like English and French, in which flexion is reduced. This effect is obvious as strong flexion generates many extra words that reduce the ratio of edges to nodes in the related networks.

Second, a different situation is seen for small networks with $10 < N < 1000$ ($N$ represents the number of unique words, so the corresponding actual text lengths can be substantially larger than this). Such networks are emanation of text fragments with the length ranging from a printed line to a chapter. These lengths are by far insufficient to reflect the global statistical properties of language, but $-$ apart from the statistical fluctuations of using the words, which manifest themselves here $-$ this is exactly these lengths that are the principal carriers of both the grammatical rules and author styles. Generally, they describe local properties of texts and language, and a significant part of language complexity is encoded just on this level. Each panel of Fig.~\ref{fig.texts.aspl} shows ASPL calculated for growing a different piece of a novel. The variability among the maximum magnitudes of $L(N)$ for different pieces of the same novel illustrate the statistical fluctuations, but not all the observed differences are equally trivial. For example, let us look at the plots obtained for \textit{Ulysses} (panel (a)). The curves representing ASPLs for different pieces of this work show not only the different maximum heights, but also their variable widths and locations. Such changing behaviour of ASPL is observed for no other novel shown. However, what \textit{Ulysses} is known for is its unequal stylistic heterogeneity: each chapter represents different style, literary epoch, or even literary genre. This suggests that the results for ASPL in this case reflect just this heterogeneity. Other notable indications of the style influence on ASPL are \textit{The Voyage of the Beagle} by C.~Darwin, the only non-literary work in our set, with its largest maximum height seen in Fig.~\ref{fig.languages.aspl}(a), \textit{Pan Tadeusz} by A.~Mickiewicz (Fig.~\ref{fig.languages.aspl}(c)) $-$ the only poem, and \textit{Memoires d'Hadrien} by M.~Yourcenar, which uses peculiar style.

\begin{figure}
\hspace{-0.5cm}
\includegraphics[scale=0.26]{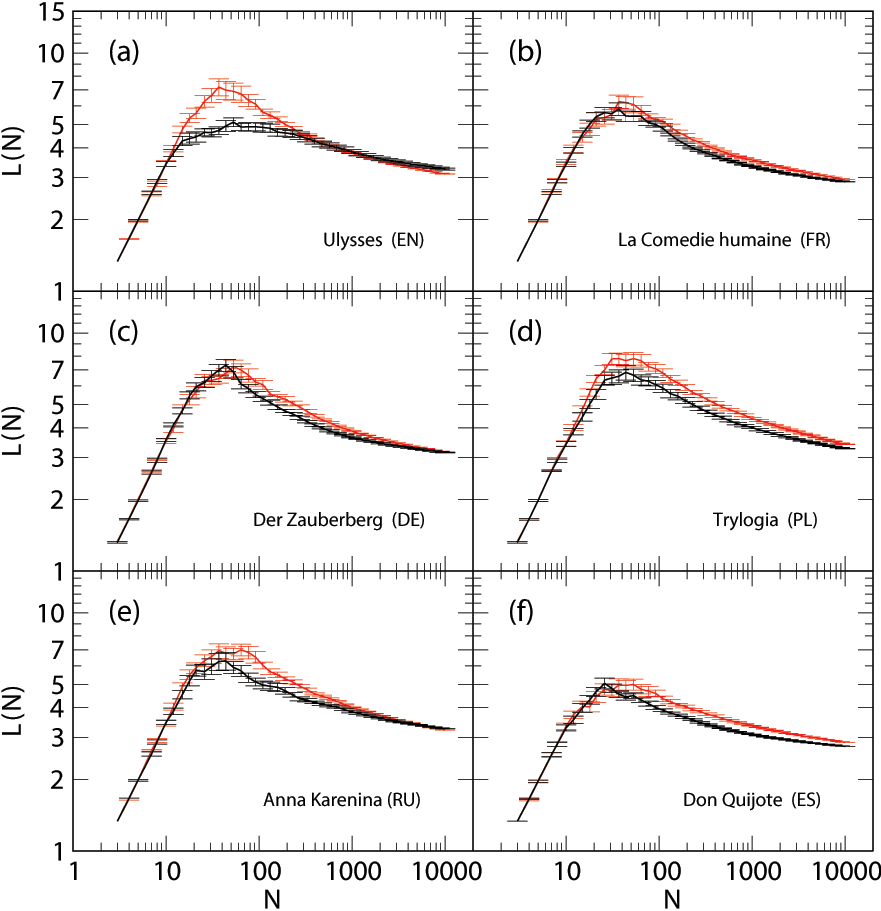}
\caption{(Color online) Evolution of the average shortest path length $L(N)$ for growing word-adjacency networks created from the original (black) and randomly reshuffled (red) literary texts (the same ones as in Fig.~\ref{fig.texts.aspl}). The curves representing $L(N)$ for the original texts were averaged over 22-30 different pieces of each text (depending on a text's length), while for the surrogate texts they were averaged over 20 independent realizations of text reshuffling. Error bars denote standard error of the mean. Note that qualitatively similar shape of the curves for both kinds of data is by no means surprising: the word-frequency statistics is one of the key factors that exert influence on topology of the word-adjacency networks and this statistics  is invariant under text randomization.}
\label{fig.texts.reshuffled.aspl}
\end{figure}

How literary styles or authors' own writing styles influence statistical properties of the corresponding networks is rather a complex issue that is beyond the purpose of the present work, but it seems doubtless that stylistic fingerprints have a tendency to manifest themselves in local network structures, while the global structure is largely style-free. This may be viewed as a parallel phenomenon to the one known from the Zipfian analysis of word frequencies, where the global power-law form of the corresponding rank distributions has rather a universal character, while the particular authors' styles are encoded locally, primarily in attributing specific ranks to specific words.

In the present context, we may say that the stylistic fingerprints can have two main appearances: the above-mentioned specific word ranks and the local correlations in word occurrences. While the former is difficult to be reflected in the results of our statistical analysis, the latter can to some extent be seen in the behaviour of ASPL. Let us notice that local correlations can be context-related and can lead to local distrotion of the overall word-usage frequencies, e.g., by increasing the frequency of certain words. From the network perspective, an increased frequency of any word causes shortening of ASPL in respect to an uncorrelated text. Thus, one may expect that the original texts show shorter ASPLs than the surrogate texts obtained by randomly shuffling the order of words. Magnitude of this effect and the range of network size in which it is observed may be different for different texts, but its existence should be universal. Indeed, Fig.~\ref{fig.texts.reshuffled.aspl} exhibits that ASPL increases after reshuffling for each considered text and that the magnitude of the difference is text-dependent. This effect is the most pronounced for \textit{Ulysses} and the weakest for \textit{La Com\'edie humaine}. The error bars reflect the statistical significance of the result. What is equally important is that any difference between the related ASPLs asymptotically decreases with increasing $N$ (except for \textit{Ulysses}), which supports our statement that the large-scale network properties are more universal than the local-scale ones.

The exception of \textit{Ulysses}, where the surrogates' $L(N)$ falls below the original one for $N > 2000$, reflects another effect that acts opposite to the correlation-based one. Actually, it is an artifact of our definition of the network edges: we use binary edges irrespective of how often a given word pair occurs in a piece of text. Therefore, in any original text the number of edges is smaller than the number of actual word-neighbour pairs, while in a surrogate text, owing to destruction of such repeated 2-grams, the number of edges increases, which obviously leads to $L(N)$ decrease. In \textit{Ulysses} there are more frequent 2-grams than in typical literary texts and this can account for the observed peculiarity.

Taking all our results into consideration, it should be underlined that the observed shape of ASPL cannot be treated purely as a statistically meaningless effect of a small sample. In spite of this it often happens that a network model is considered appropriate if it reproduces some properties of real data only in the limit of large samples. Here we definitely cannot follow this path: as indicated above, in natural language some of the most important features like grammar and style may shape the corresponding network's topology on local scale causing it being substantially different than the global topology. This is why we believe that a satisfactory linguistic model has to reproduce both the local and the global properties of empirical data.

\section{Modelling linguistic networks}

\subsection{The Dorogovtsev-Mendes model}
\label{sec.dm.model}

A relation between the accelerated growth of a network and the growth of a piece of text was noticed soon after such networks were introduced, so their linguistic applications have already a long record~\cite{dorogovtsev2001b}. Our objective is to inspect, how well the accelerated-growth network models can reproduce the ASPL shape known from the empirical data. We start with the generic model introduced by Dorogovtsev and Mendes (the DM model henceforth)~\cite{dorogovtsev2000a}.

Let us consider a network of size $N(t)=t+n_0$, i.e., such that the growth starts from an initial ``network seed'' with $n_0$ nodes interconnected by $e_0 \ge n_0-1$ edges, and at each time step $t$ a new node is added. This node connects itself to $m$ existing nodes by undirected, binary edges with probabilities $\pi(i) \sim k_i$, where $k_i$ stands for the $i$th node's connectivity. At the same time, $c(t)$ new edges are formed among the existing nodes in such a way that neither multiple nor self-looping edges are allowed and the probability of connecting the nodes $i,j$ is defined by $\pi(i,j) \sim k_i k_j$. Thus, at each moment the network grows by one node and $m+c(t)$ edges. In general, $c(t)$ can have either discrete or continuous values, but it is convenient to assume the latter. In the trivial case $m > 0$ and $c(t)=0$ and the network grows according to the Barab\'asi-Albert (BA) model of pure preferential attachment with no edges formed inside the network. As it is well known, this scheme leads to a scale-free connectivity distribution $P(k) \sim k^{-\gamma}$ with $\gamma = 3$. Although satisfactorily describing some real-world systems, this model cannot be applied to the word-adjacency networks, because it does not reflect the real procedure of text creation.

The accelerated growth can be attained by a monotonously increasing function $c(t)$. Let it be a power function of time:
\begin{equation}
c(t)=c_0 t^{\alpha}
\label{eq.continuous}
\end{equation}
with $c_0 > 0$ and $\alpha > 0$. The so-defined network consists of the same number of nodes as before, i.e., $N(t)=t+n_0$, but now the number of edges increases in a nonlinear way:
\begin{equation}
E(t) = mt + {c_0 \over \alpha+1} t^{\alpha+1} + e_0.
\label{eq.acceler.growth.edges}
\end{equation}
Note that if $\alpha > 1$, the fully connected state is an attractor and it will be reached in finite time. From the linguistic perspective, however, such a state, in which every word neighbours all other words, is forbidden by grammar. Thus, in realistic approach $\alpha < 1$ or, if $c_0 \ll 1$ and one considers networks of limited size not exceeding its empirical values, $\alpha \le 1 + \epsilon$ with $\epsilon \ll 1$.

In simulations, the continuous character of $c(t)$ can be approximated by expressing it by a sum of two terms: $c(t) = c_{\rm int}(t) + p(t)$, where $c_{\rm int}(t)$ is the integer part of $c(t)$. Then at each time step $c_{\rm int}(t)$ edges are added to the network with probability 1 and an additional edge $-$ with probability $p(t)$ if $p(t) > 0$.

It is worthwhile to notice that the exponent $\alpha$ in Eq.~(\ref{eq.continuous}) is related to the Heaps exponent $\delta$ in Eq.~(\ref{eq.heaps}). First, note that a step of $t$ in Eq.~(\ref{eq.continuous}) corresponds to adding a new node to the network (i.e., writing a new word), while a step of $\tau$ in Eq.~(\ref{eq.heaps}) corresponds to adding a new edge (i.e., writing \textit{any} word). Now, if one neglects multiple occurrences of the same word pairs in text, $\tau$ is equal to the total number of edges $E(t)$ in Eq.~(\ref{eq.acceler.growth.edges}). In parallel, $N(\tau)$ can be identified with $t$. This means that in this case the Heaps law may be expressed by $t \sim [E(t)]^{\delta}$, which gives: $E(t) \sim t^{1/\delta}$. Thus, for sufficiently large $t$ one may neglect $mt+e_0$ in Eq.~(\ref{eq.acceler.growth.edges}) and arrive at the following relation:
\begin{equation}
\alpha = {1 \over \delta} - 1 .
\label{alpha.delta}
\end{equation}
This relation implies that reasonable values of $\alpha$ are determined by the empirical Heaps exponents and fall in the range between $\alpha \approx 0.1$ for $\delta=0.9$ and $\alpha=1.0$ for $\delta=0.5$. Actually, in empirical data the Heaps law is not valid for the whole range of $\tau$ (see, e.g.,~\cite{petersen2012}) and both the exponents $\alpha$ and $\delta$ are functions of $t$.

As a side remark, it is interesting to notice that shape of the declining phase of $L(N)$ for the empirical data in Fig.~\ref{fig.texts.aspl} can easily be approximated by a simple function that can be derived from an assumption that the word-adjacency networks show both the features of the classical ER graphs and the graphs with accelerated growth. ASPL for ER graphs is given by $L(N) \sim \ln N / \ln \langle k \rangle$, where $\langle \cdot \rangle$ denotes the average~\cite{albert2002}. Then, by substitution of $\langle k \rangle = 2 E(t)/N(t)$, where $N(t) \approx t$, after some algebra one arrives at the following relation valid for sufficiently large $N$:
\begin{equation}
L(N) \sim {\ln N \over \ln {c_0 \over \alpha + 1} + \alpha \ln N}.
\end{equation}
This form of $L(N)$ may be fitted to its empirical values and one can obtain acceptable agreement between both (not shown). Obviously, the word-adjacency networks are not of the ER type, but nevertheless they can possess some features that allow for such rough approximation. We do not discuss this analogy further, though.

\begin{figure}
\includegraphics[scale=0.32]{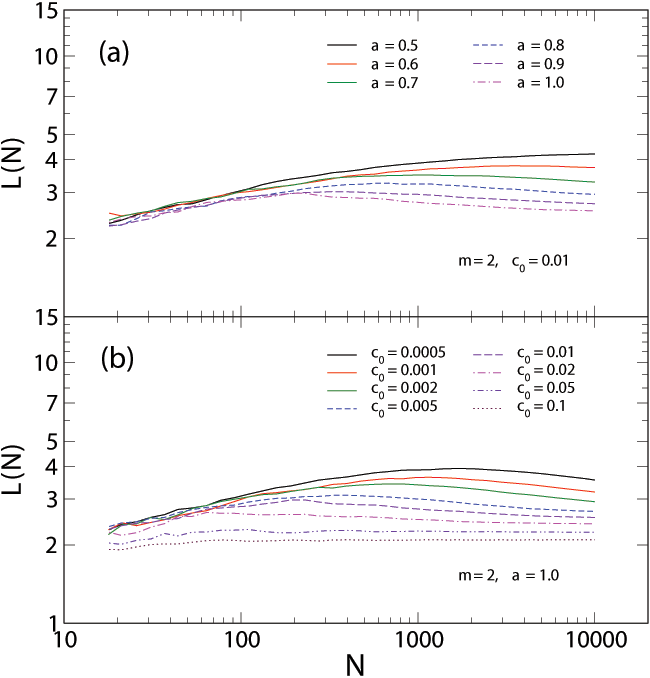}
\caption{(Color online) Evolution of the average shortest path length $L(N)$ for networks with accelerated growth simulated according to the Dorogovtsev-Mendes model. The networks are characterized by three parameters: $m, \alpha, c_0$ (see Eq.~(\ref{eq.acceler.growth.edges})), one of which is varied ($\alpha$ in (a) and $c_0$ in (b)), while the other two are fixed. In both panels vertical axes have the same range as in Fig.~\ref{fig.texts.aspl} and Fig.~\ref{fig.languages.aspl} in order to faciliate comparison. Only the results for $N>15$ are shown, because for smaller $N$ the particular choice of initial condition as a chain of $n_0=7$ nodes distorts the behaviour of ASPL. The legends describe the lines as they appear from top ($\alpha=0.5$ or $c_0 = 0.0005$) to bottom ($\alpha=1.0$ or $c_0 = 0.1$).}
\label{fig.aspl.dm}
\end{figure}

The plots in Figure~\ref{fig.aspl.dm} present $L(N)$ for the accelerated-growth networks constructed according to the DM model with different values of the parameters $c_0$ and $\alpha$. The third parameter was fixed at $m=2$, because this value approximates the topology of real texts the best. Each new word immediately receives there two direct neighbours: a preceding and a subsequent word. (No word in any piece of text has the connectivity $k=1$ except for the first and the last ones, and even in this case this is so only if these words are used once in the whole text, which happens rarely). As one might expect, for a given value of $N$, the stronger is the acceleration expressed by $\alpha$, the more intra-network edges appear at each step and the shorter is ASPL. Obviously, the same refers to the ASPL dependence on the parameter $c_0$, which is effectively responsible for how early the acceleration mechanism enters the network growth. As regards the ASPL dependence on $N$, for small networks with several tens of nodes, if $\alpha$ and $c_0$ are also small, the acceleration has not started yet and the network grows according to the standard preferential attachment with increasing $L(N)$. Then the acceleration is switched on for some $N$ and $L(N)$ starts to decrease owing to appearance of the intra-network edges. This leads to formation of a maximum of ASPL. On the other hand, if $c_0$ and $\alpha$ are sufficiently large, the shortening of ASPL due to acceleration is present since the very beginning and no maximum has a chance to form.

If one compares the behaviour of $L(N)$ observed for the empirical networks (Figs.~\ref{fig.texts.aspl} and~\ref{fig.languages.aspl}) with that obtained from simulations (Fig.~\ref{fig.aspl.dm}), a strong discrepancy is evident for $N<10^4$. The simulated networks are much more condensed and unable to develop sufficiently high initial values of $L(N)$ for any of the possible parameter combinations. Actually, this is achievable, but only for a specific choice of the initial condition, in which the network seed forms a long chain consisting of a number of serially connected nodes. In this case $L(N)$ can be escalated up to any conceivable level and then, obviously, it can only decrease to more standard values, which leads to $L(N)$ that is more or less concordant with the empirics. However, we ignore this case being both trivial and topologically inappropriate.

\begin{figure}
\includegraphics[scale=0.15]{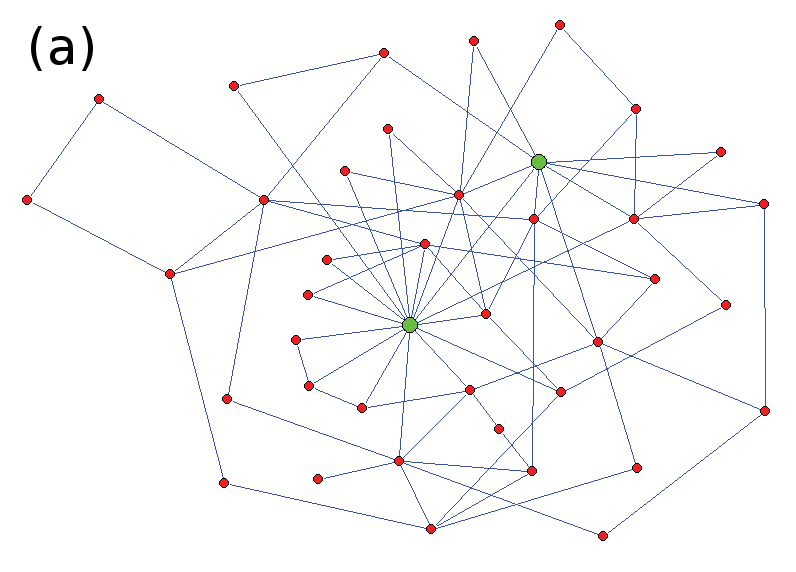}
\includegraphics[scale=0.15]{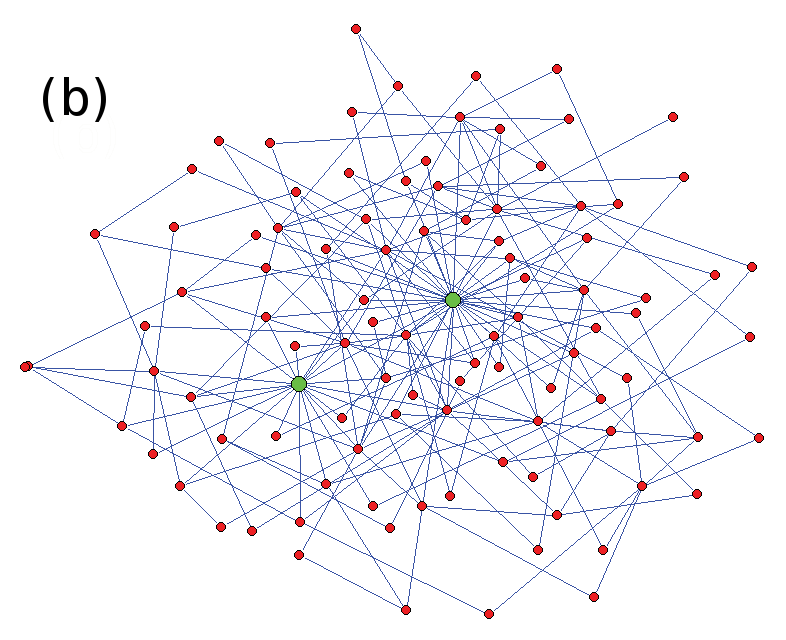}

\includegraphics[scale=0.15]{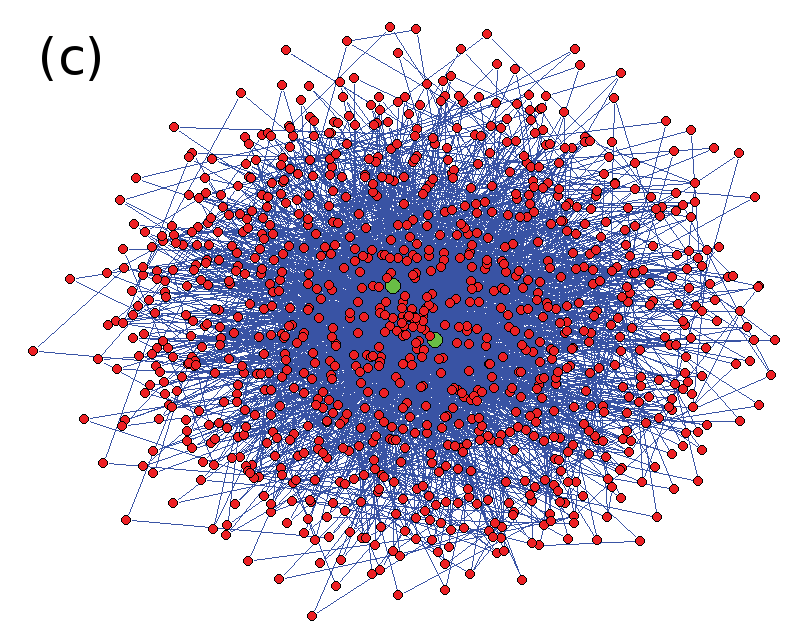}
\caption{(Color online) Visualisation of a typical network constructed according to the DM model ($c_0 = 0.01$, $\alpha = 0.8$, $m=2$, with a network seed being a chain of $n_0=7$ nodes). Three snapshots of the network development are shown with the same values of $N$ as in Fig.~\ref{fig.networks}: $N=40$ (a), 100 (b), and 1000 (c).}
\label{fig.dm.networks}
\end{figure}

A reason for which the DM model does not work well for ASPL in the case of the word-adjacency networks can be learned from visualizations shown in Fig.~\ref{fig.dm.networks}. The presented graphs constitute three snapshots of the same simulated DM network corresponding to different values of $N$. What can imediately be noticed is the complete absence of the long loops of serially connected nodes that are so characteristic for empirical data (compare Fig.~\ref{fig.networks}). These loops are products of writing the passages of text that consist solely of words not used before. While this is the most frequent case in the beginning stages of text creation, such pure sets of novel words can be found at any later stage as well, though they are then less frequent and many of the earlier formed loops are already destroyed by the long-distance edges. Nevertheless, the loops are an important property of the empirical networks' topology and, as such, their existence cannot be ignored in designing a model. Unfortunately, the DM model was applied to language merely as a minimal tool of modelling its large-scale properties like the connectivity distributions $P(k)$ or the asymptotic behaviour of ASPL, and therefore it ignores the local aspects of network topology. However, what else the model does satisfactorily, is reproducing the early emergence of local hubs, which are seen even for small empirical networks with $N < 100$ (Fig.~\ref{fig.dm.networks}(a,b)). Here we find a good agreement with the data.

Our findings that the DM model cannot reproduce important properties of linguistic networks go in parallel with the results reported in Ref.~\cite{masucci2006}, according to which the DM model misses also the empirical values of the average clustering coefficient obtained from the word-adjacency networks. The authors of that Ref. managed to improve performance of the model by extending it to incorporate different mechanisms of attachment, both the preferential and random ones. However, while those new mechanisms offer some interesting modifications that are in the spirit of real language (like, e.g., reducing the probability of creating triangles among the nodes), from the point of view of ASPL they are still not satisfying, because they do not create the long loops of nodes. The same can be said about other extensions of the DM model discussed earlier in the linguistic context, like the model with edge rewiring. We considered both its original static form with a constant number $r$ of rewired edges in each step and its modified version with variable $r(t)$ being a power function of $t$, in analogy to $c(t)$ (Eq.~(\ref{eq.continuous})). However, both versions generate networks that qualitatively resemble the ones in the generic model, so neither of them is satisfactory from the ASPL perspective.

\subsection{Alternative models}
\label{sec.alternatives}

\subsubsection{Network Simon-Heaps model}

In order to find a mechanism that can generate networks with the chain-like loops, but also with hubs and short ASPL, we focus our attention on stochastic processes that resemble writing text. One of the simplest processes of this kind is the Simon process, known for its Zipf-like statistics of values and the preferential attachment paradigm~\cite{simon1955}. According to the Simon algorithm, new words are added to a piece of text with constant probability $p_0<1$, while the already-used words are added with probability proportional to their frequency in the written part of text. However, this algorithm leads to the equilibrium growth, so it cannot reproduce texts that fulfill the Heaps law, for example. A more realistic case can be obtained if we allow for variable probability $p(\tau)$ of adding new words, where $\tau$ is the current length of text. Its functional form can be derived from the Heaps law (Eq.~(\ref{eq.heaps})):
\begin{equation}
p(\tau) \sim \tau^{\delta-1}
\label{eq.simon}
\end{equation}
with $\delta < 1$. Like in real texts, as $\tau$ grows, the probability of using new words decreases. In particular, for every new word written, $K(\tau) = 1/p(\tau)-1$ old words have to be written as well, with $K(\tau)$ increasing. From the network perspective, this is equivalent to adding $K(\tau)$ intra-network edges for every new node connected to the network. The resulting network may thus be counted among the networks with accelerated growth, but not of the DM type.

Purely on the network level, the Simon-Heaps (SH) algorithm can be realized as follows. Let us start with a network seed consisting of a single node or a group of $n_0$ connected nodes forming, e.g., a chain. In the first step of the algorithm, a new node is connected to one of the seed nodes chosen at random. In each of the subsequent steps $t$, the network can be grown by adding strictly one new node or one new edge with the probability $p(t)$ and $1-p(t)$, respectively. If this is a new node, it has to be connected to the latest involved node (by the involved node we mean: (1) the one that was added in the step $t-1$, or (2) the one that was connected by an edge in the step $t-1$ and that \textit{was not} added in the step $t-2$). If a new edge is added, one of its ends has to be attached to the latest involved node, while the other end has to be attached to a node $i$ chosen preferentially according to its degree $k_i$. We also imply a restriction that no edge may be doubled. This algorithm can be viewed as a kind of preferential random walk on a set of initially disconnected nodes with each step creating an edge.

\begin{figure}
\includegraphics[scale=0.15]{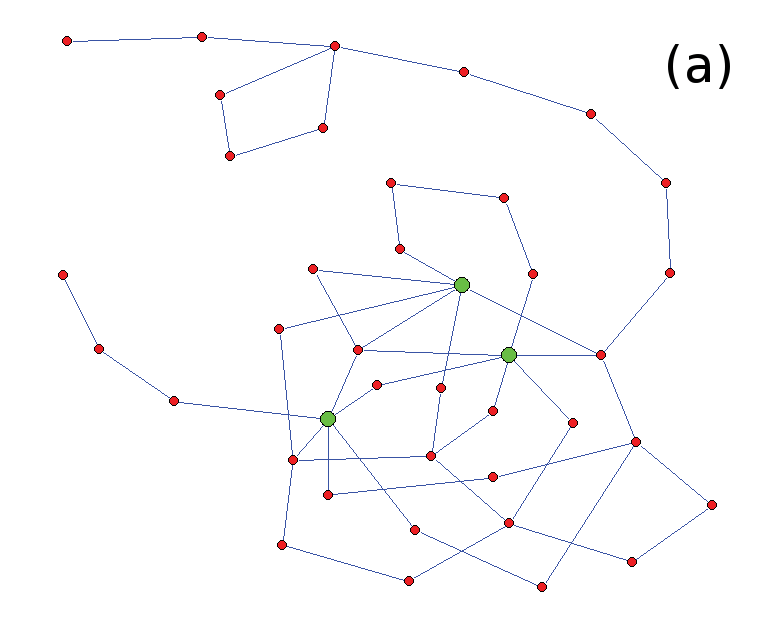}
\includegraphics[scale=0.15]{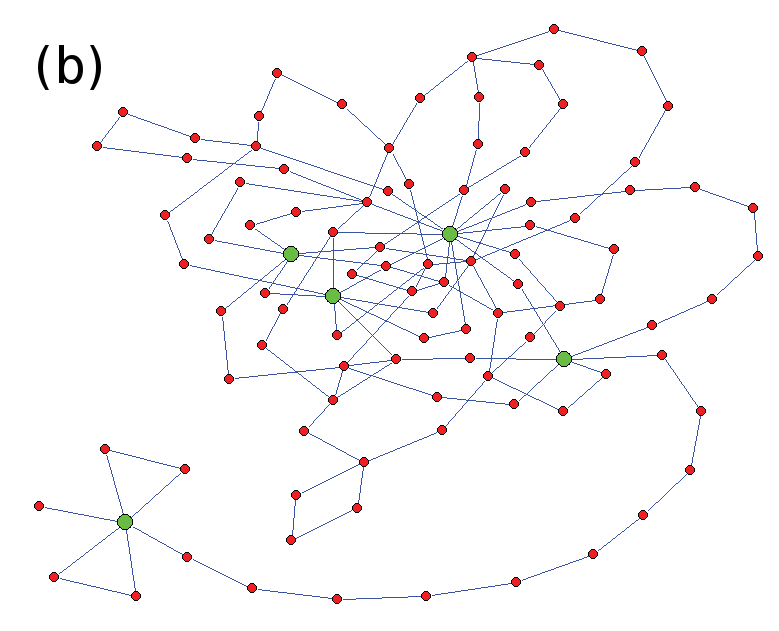}

\includegraphics[scale=0.15]{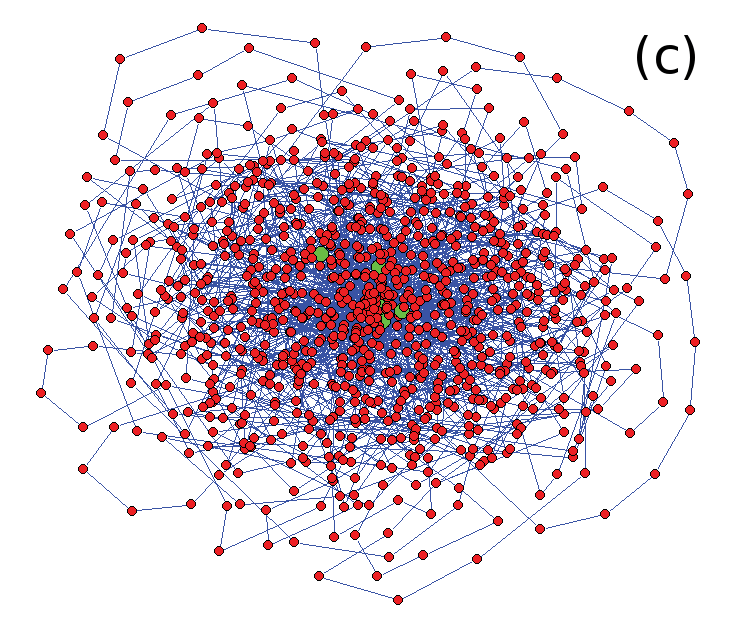}
\caption{(Color online) Visualisation of a network constructed according to the Simon-Heaps model with $p_0 = 1.0$ and $\mu = 0.075$. Three snapshots of the network development are shown: $N=40$ (a), 100 (b), and 1000 (c). The earliest developed hubs are distinguished by larger symbols (green/light gray).}
\label{fig.sh.networks}
\end{figure}

\begin{figure}
\includegraphics[scale=0.3]{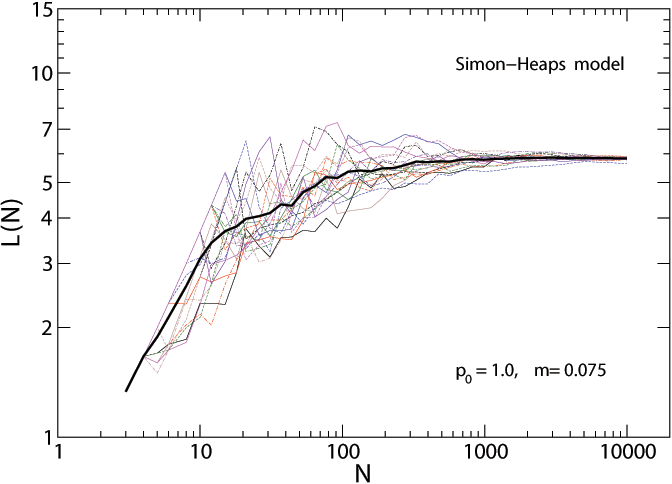}
\caption{(Color online) Different realisations of the average shortest path length $L(N)$ for sample networks constructed according to the Simon-Heaps algorithm with $p_0=1.0$ and $\mu = 0.075$ (thin lines) together with the respective average of all the individual resalisations (thick line).}
\label{fig.sh.aspl}
\end{figure}

Eq.~(\ref{eq.simon}) implies that the time-dependent probability $p(t)$ of adding a new node is a decreasing power function of $t$:
\begin{equation}
p(t) = p_0 t^{-\mu} ,
\label{eq.sh.probability}
\end{equation}
where $\mu > 0$ that guarantees the accelerated growth. As at the beginning new words have to dominate the network's growth, a choice of $p_0 = 1$ is justified, which effectively gives us a very simple, one-parameter model. For small values of $\mu$ ($\mu \ll 1$) the growth is realised principally by creation of long loops of nodes with $k=2$, while the hubs are numerous and of moderate degree. On the other hand, the larger is $\mu$, the faster is the $p(t)$ decrease and for $\mu \approx 1$ the structure is based on 1-2 hubs of extremely high centrality and it almost completely lacks significant loops. The optimum structure can thus be found somewhere in between these two extremes $-$ an example is shown in Fig.~\ref{fig.sh.networks}. If compared with Figs.~\ref{fig.networks} and~\ref{fig.dm.networks}, one can see that the SH model produces networks whose visual structure resembles more the empirical ones than the generic DM networks do. However, we still do not observe a satisfying agreement. This impression receives strong quantitative support from Fig.~\ref{fig.sh.aspl}, where ASPL for the SH networks with the same choice of $\mu=0.075$ as in Fig.~\ref{fig.networks} is exhibited. Its dependence on the network size $N$ does not show any maximum and it is, on average, a monotonous increasing function up to $N \approx 10^4$, where it saturates. By varying a value of $\mu$, we observe a related variation of the saturation level, but there is no qualitative change in the overall ASPL behaviour (not shown).

\begin{figure}
\includegraphics[scale=0.3]{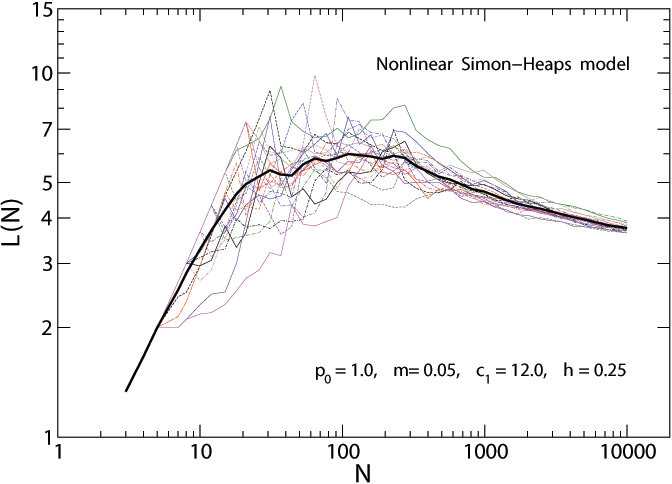}
\caption{(Color online) Average shortest path length $L(N)$ for sample networks constructed according to the Simon-Heaps model with the parameters $p_0 = 1.0$ and $\mu = 0.05$ and with nonlinear preference expressed by the parameters $c_1 = 12.0$ and $\eta = 0.25$. Different realisations of the model are denoted by thin lines, while the average by thick line.}
\label{fig.sh.nonlin.aspl}
\end{figure}

\begin{figure}
\includegraphics[scale=0.15]{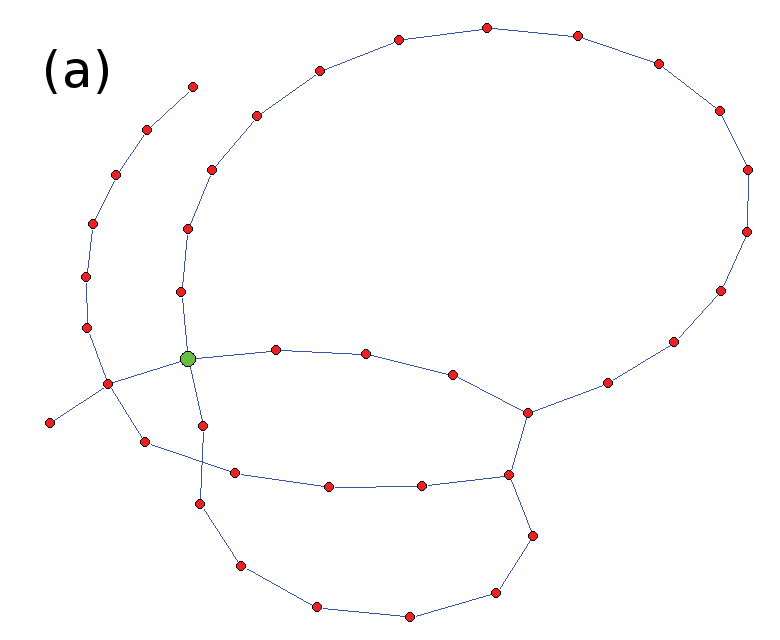}
\includegraphics[scale=0.15]{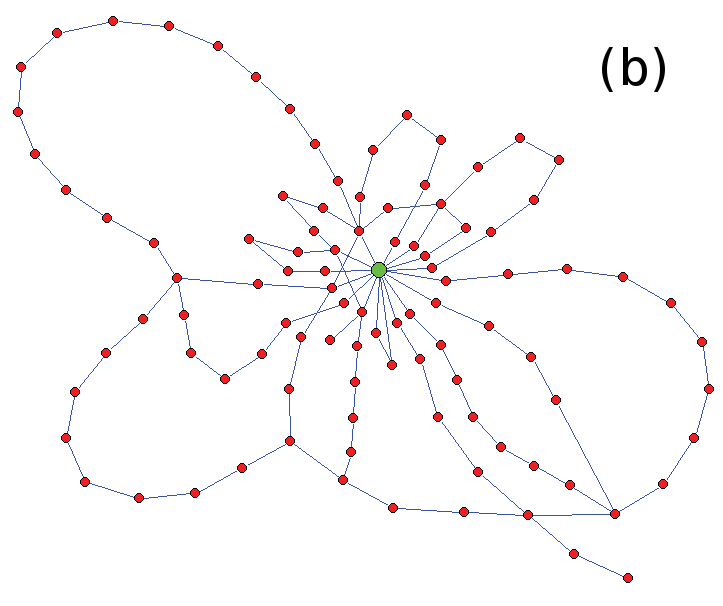}

\includegraphics[scale=0.15]{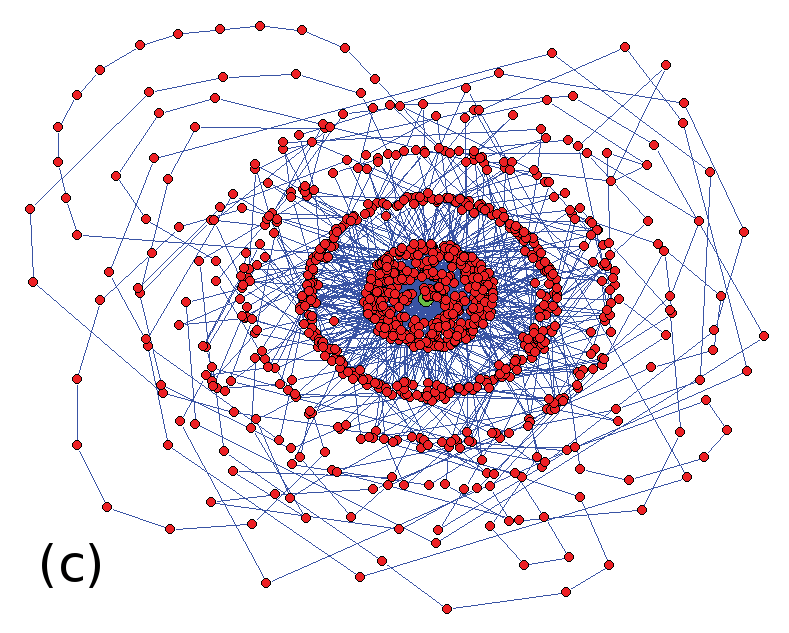}
\caption{(Color online) Visualisation of a network constructed according to the nonlinear Simon-Heaps model with the same parameter values as in Fig.~\ref{fig.sh.nonlin.aspl}. Three snapshots of the network development are shown: $N=40$ (a), 100 (b), and 1000 (c). The earliest developed hubs are distinguished by larger symbols (green/light gray).}
\label{fig.sh.nonlin.networks}
\end{figure}

This suggests that a one-parameter model is insufficient. What we therefore need in a more realistic model, is to secure that the resulting networks in their juvenile stage show both the loops and the well-developed hubs $-$ a requirement that was impossible to be met in the one-parameter model. This can be achieved by amplifying connectivity of the hubs via nonlinear preferential attachment, while leaving $\mu$ in Eq.~(\ref{eq.sh.probability}) as it is. The amplification should be in action mainly for small $t$, so we postulate the nonlinear preference to be of the following, double-power form:
\begin{equation}
\pi(k) \sim k^{\xi(t)}, \quad\qquad \xi(t) = c_1 t^{-\eta} ,
\label{eq.nonlin.pref}
\end{equation}
where $c_1 > 0$ and $\eta > 0$. This form assures that as the network grows, the preferential attachment rule becomes closer and closer to the standard, linear one. Now, the so-defined nonlinear version of the SH model comprises three parameters: $\mu$, $c_1$, $\eta$ and works much better as regards the evolution of ASPL. In Fig.~\ref{fig.sh.nonlin.aspl} we present $L(N)$ for this model with a particular choice of parameters that gives ASPL that qualitatively reproduces its empirical behaviour for such text as \textit{Ulysses} (Fig.~\ref{fig.texts.aspl}a) $-$ something that can be achieved neither by the DM model nor by the linear SH model. Despite the fact that if we look at the early stages of the network growth displayed in Fig.~\ref{fig.sh.nonlin.networks}ab, some similarity between the empirical and the model networks can be pointed out, the main disappointment from the model comes from the emergence of a hub with unrealistically high centrality, which gradually dominates the whole structure (Fig.~\ref{fig.sh.nonlin.networks}c). Obviously, this failure eliminates the nonlinear SH model in its current form from our further consideration. However, in the forthcoming we do not abandon the idea of attaching new nodes to the latest involved nodes as we view it being realistic.

\subsubsection{Hybrid model with acceleration and chain growth}

The preceding discussion has shown that both variants of the networks with accelerated growth, i.e. the DM models (Sect.~\ref{sec.dm.model}) and the SH models (Sect.~\ref{sec.alternatives}), have significant drawbacks that make them rather insufficient as potential tools of modelling the word-adjacency networks. Nevertheless, these models have also advantages that are worth preserving.

We therefore propose a hybrid model that goes in this direction by inclusion of two distinct regimes of adding new nodes: a new node can be attached to the latest involved node, like in the SH models or, alternatively, it can be attached to one or more old nodes via linear preferential attachment, like in the generic DM model. The first regime is responsible for forming locally the chain-like loops, while the second regime is responsible for forming the large-scale structure. Switching between these regimes is probabilistic with time-dependent probability $p(t)$ that the chain regime is chosen in a step $t$ and probability $1-p(t)$ that the accelerated-growth (DM) regime is chosen. $p(t)$ should be defined as a monotonous decreasing function of $t$ in order to be in agreement with the empirical data, where the probability of forming long loops decreases with the network size. As in the case of the SH model (Eq.~(\ref{eq.sh.probability})), we propose a power function:
\begin{equation}
p(t) = p_0 t^{-\mu} ,
\end{equation}
where $p_0 \approx 1$, $\mu > 0$, and $\mu \ll 1$ for a slow decay of $p(t)$. 

\begin{figure}
\includegraphics[scale=0.3]{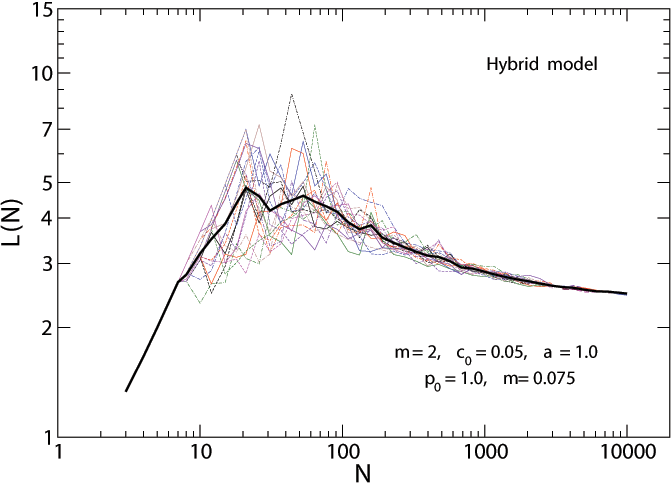}
\caption{(Color online) Average shortest path length $L(N)$ for networks constructed according to the hybrid model with $m=2$, $c_0=0.05$, $\alpha=1.0$, $p_0=1.0$, and $\mu=0.075$. Different realisations are represented by thin lines, while the average by thick one.}
\label{fig.hybrid.aspl}
\end{figure}

\begin{figure}
\includegraphics[scale=0.14]{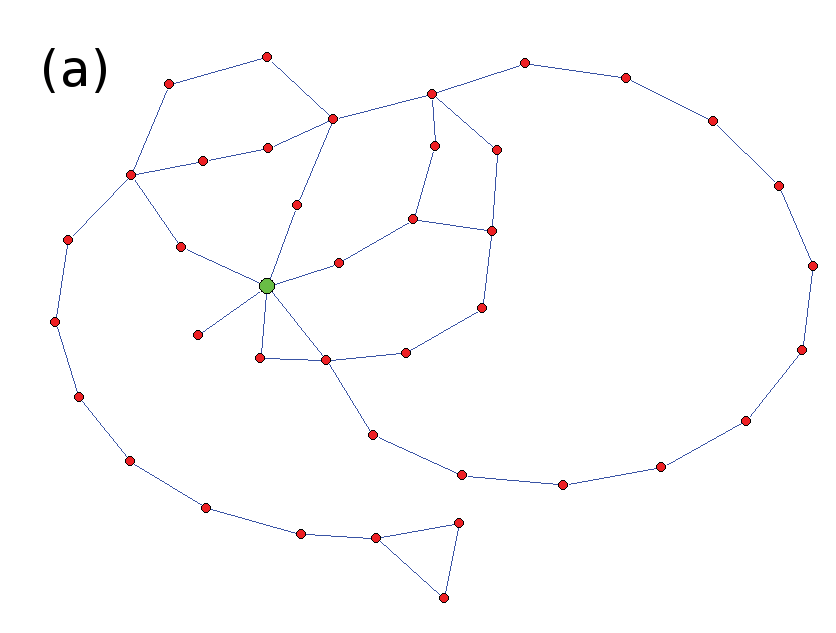}
\includegraphics[scale=0.14]{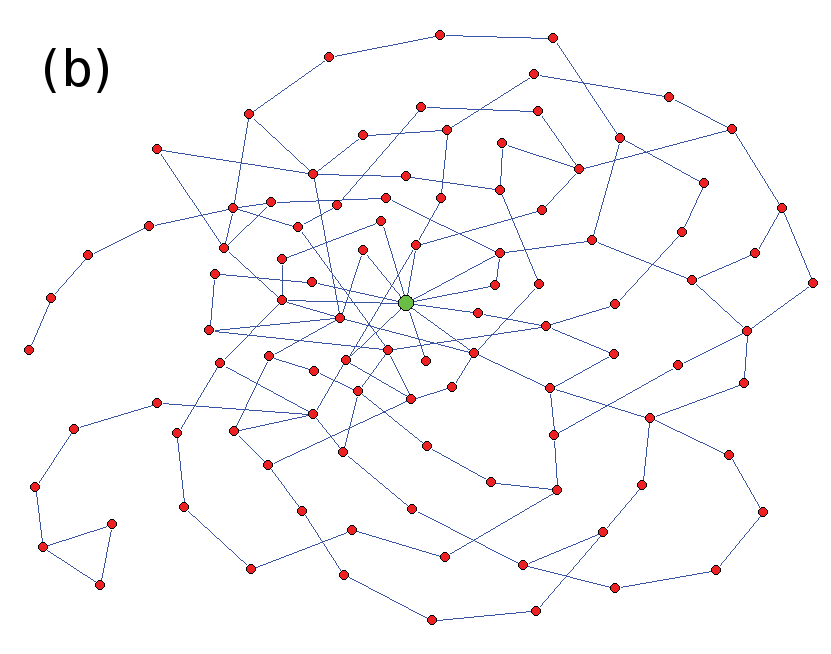}

\includegraphics[scale=0.14]{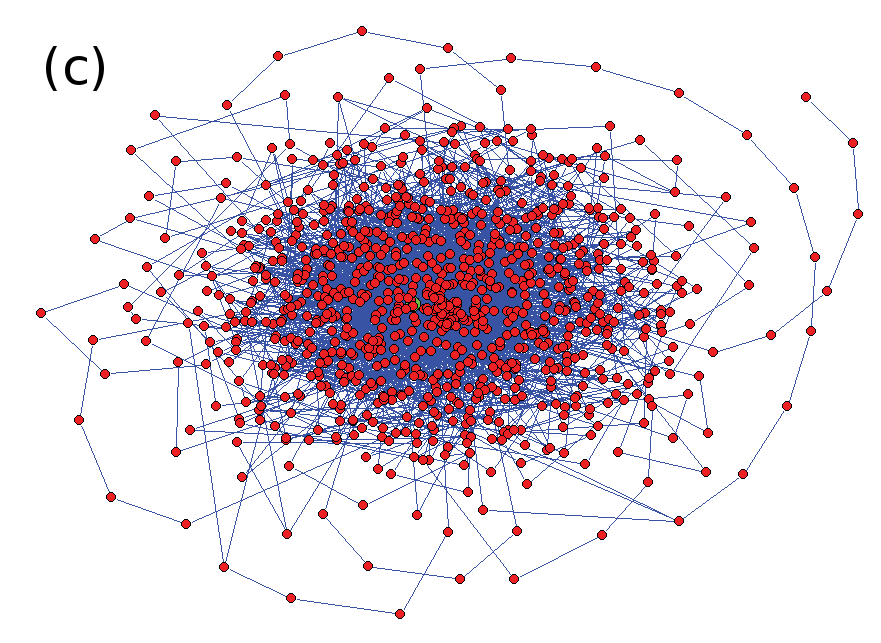}
\caption{(Color online) Visualisation of a network constructed according to the hybrid model ($c_0 = 0.01$, $\alpha = 1.0$, $m=2$, $\mu=0.075$). Three snapshots of the network development are shown with the same values of $N$ as in Fig.~\ref{fig.networks}: $N=40$ (a), 100 (b), and 1000 (c). The earliest developed hubs are distinguished by larger symbols (green/light gray).}
\label{fig.hybrid.networks}
\end{figure}

In the accelerated growth regime, the network behaves in almost exactly the same manner as in the generic DM model (or one of its extensions, if needed). The only difference is that the first new node that appears after the regime switching is obligatorily connected by one of its edges to the previously added node, in order to close the formed loop. This means that only the remaining $m-1$ edges of such a node may be connected preferentially. In the case of $m=1$ such a loop remains open unless one of the new intra-network edges closes it. From the ASPL perspective, the accelerated-growth regime reassures that this quantity shows a monotonous decline with the network size $N$ (for the adequate values of the parameters), while the chain regime can drive ASPL high for small $N$. This dual behaviour can be seen in Fig.~\ref{fig.hybrid.aspl}, indeed. What is especially welcome is that the average evolution of $L(N)$ is now able to mimic the one for real language (see Figs.~\ref{fig.texts.aspl} and~\ref{fig.languages.aspl}) to an even better degree than the nonlinear SH model does. Moreover, the structure of a corresponding sample network of size $N=1000$, visualised in Fig.~\ref{fig.hybrid.networks}c, displays no sign of the strong centrality seen in Fig.~\ref{fig.sh.nonlin.networks} for the nonlinear SH model. We feel that this structure seems the most realistic among the models considered in this work.

Of course, we do not consider this hybrid model perfect. First of all, we restricted our analysis to ASPL and to visual inspection of the simulated networks. How the proposed models perform themselves from a point of view of other network measures like, for example, the clustering coefficient or the degree distribution of nodes, is beyond the scope of this analysis. Second, Fig.~\ref{fig.hybrid.networks}ab suggests that the model demands further improvements regarding the early structure of networks, in which the hubs are fewer and less evident among the nodes than in the empirical networks, as this property might influence the later stages of the network growth. So it is conceivable, that an even more realistic model should contain also such growth rules as the local preferential attachment or the attachemnt of nodes to predefined hubs, as in the model proposed in~\cite{masucci2006}.

\section{Conclusions}

Natural language is a complex system and like all other systems of this kind it develops important features at all scales of its organization~\cite{kwapien2012}. No scale may be neglected as being meaningless from the point of view of a correct description and modelling. This is true for the whole spectrum of scales, from the scale of letters or phonemes, where the phonetic properties are expressed, through the scale of words and phrases, where the key role is played by grammar and style, up to the scale of large national corpora involving tens of thousands of text samples, at which the most global statistical properties of language are manifested. Thus, any approach, in which only the large-scale (or even asymptotic) statistical properties are of interest, seems to be by far insufficient. An example of such an approach is modelling of empirical word-adjacency networks by the networks with accelerated growth~\cite{dorogovtsev2001b,markosova2008}. While these models offer results that agree with the empirical ones for the word-adjacency networks of large size, they perform worse in describing local network topology~\cite{masucci2006}, which encodes some important properties of language.

In our work we studied properties of the word-adjacency networks constructued from literary texts written in different European languages. We observed growth of these networks representing a process of text creation. We focused our attention on the evolution of the average shortest path length $L$ as a function of the network size $N$. We found that ASPL is not a monotonous function of $N$, but it consists of two phases: a shorter phase, in which $L(N)$ increases up to its maximum value of order of 10 (reached for $N<100$), and a longer (perhaps infinite) phase of decline, in which ASPL typically falls well below $L=5$ for $N>1000$. Such behaviour of ASPL is related to local topological properties of the empirical networks, which exhibit loops of nodes connected with each other like chains. These loops are formed from the very beginning of text by the words that were not used before, and their length slowly decreases with time as more and more words are repeated and the role of the new words diminishes.

We attempted to reproduce the empirical results regarding ASPL by simulating networks growth with the well-known Dorogovtsev-Mendes model of accelerated growth~\cite{dorogovtsev2000a}. However, we realized that this model does not offer satisfying results for small networks with $10 < N < 1000$. In particular, it cannot reproduce the maximum of ASPL. Apart from the generic DM model, we considered its extensions with preferential rewiring but none of them succeeded, either. Another type of the accelerated-growth networks that we considered here was the Simon-Heaps models with linear or nonlinear preference. These models incorporate creation of the loops by their very construction, which can produce some interesting agreement with the empirics as regards ASPL, but their topological structure observed by naked eye was nevertheless rather unrealistic.

Based on selected principles of both types of models, we also proposed a hybrid model exploiting two paradigms: the preferential-attachment growth with acceleration (after the DM family of models) and the chain-like linear growth (after the SH models). While the former is responsible for the large-scale structure of the network (large $N$), the latter is able to correctly reproduce certain aspects of the local structure (small $N$). Combining these two paradigms in one model allowed us to generate networks with
ASPL reproducing its behaviour for the empirical networks.

It is worth noticing that the hybrid model proposed in this work, although its motivation was purely linguistic, may as well be considered an alternative mechanism of network growth in abstract sense. Thus, it can perhaps be applied to model evolution of other systems whose growth resembles the one considered here. For example, we suppose that certain kinds of transportation networks, like those connected with distribution of products might in certain situations reveal similar properties, i.e., the dual, chain-like and accelerated growth. Discussing this issue in more detail is beyond the scope of the present work, however.

\end{document}